\documentclass[letterpaper, 10pt,conference]{ieeeconf}  

\IEEEoverridecommandlockouts                         
\makeatletter
\let\NAT@parse\undefined
\makeatother

\usepackage{makecell} 
\usepackage[square, comma, numbers,sort]{natbib}
\usepackage[numbers]{natbib}
\usepackage{amsmath,amssymb,amsfonts}
\usepackage{algorithmic}
\usepackage{graphicx}
\usepackage{textcomp}
\usepackage{multirow}
\usepackage{bbding} 
\usepackage{pifont}
\usepackage{booktabs}
\usepackage{balance}
\usepackage{color}
\usepackage[table,dvipsnames]{xcolor}
\usepackage{listings}
\usepackage{xcolor}
\usepackage{tabularx}
\usepackage{float}
\usepackage{xspace}
\definecolor{salmon}{rgb}{0.98, 0.50, 0.45}
\usepackage{tcolorbox}
\tcbuselibrary{listingsutf8}
\usepackage{booktabs}
\usepackage{caption}
\usepackage{algorithm}
\usepackage{algorithmic}

\usepackage[pagebackref=true,breaklinks=true,colorlinks,bookmarks=false]{hyperref}

\hypersetup{
linkcolor=BrickRed
,citecolor=Green
,filecolor=Mulberry
,urlcolor=NavyBlue
,menucolor=BrickRed
,runcolor=Mulberry
,linkbordercolor=BrickRed
,citebordercolor=Green
,filebordercolor=Mulberry
,urlbordercolor=NavyBlue
,menubordercolor=BrickRed
,runbordercolor=Mulberry
}

\newtcblisting{coloredlisting}{
    listing only,
    colframe=gray!60,
    colback=white,
    boxrule=0.5pt,
    coltitle=black,
    fonttitle=\small\itshape,
    listing options={
        basicstyle=\ttfamily\scriptsize,
        breaklines=true,
        language=Python,
        escapeinside={(*@}{@*)},
        commentstyle=\color{black!40!gray},
    },
    width=\linewidth,
}

\def\BibTeX{{\rm B\kern-.05em{\sc i\kern-.025em b}\kern-.08em
    T\kern-.1667em\lower.7ex\hbox{E}\kern-.125emX}}

\newcommand{\acronym}{SeeDo\xspace}

\begin{document}
\author{Beichen Wang$^{1*}$, Juexiao Zhang$^{1*}$, Shuwen Dong$^{1\dagger}$, Irving Fang$^{1\dagger}$, and Chen Feng\textsuperscript{1\ding{41}}
\thanks{* indicates first authors with equal contributions. $\dagger$ indicates second authors with equal contributions.}
\thanks{$^{1}$All authors are with New York University, New York, NY 11201, USA {\tt\small \{bw2716, juexiao.zhang, sd5517, irving.fang, cfeng\}@nyu.edu}}
\thanks{\ding{41} Corresponding author. The work was supported in part through NSF grants 2238968, 2322242, and 2024882, and the NYU IT High Performance Computing resources, services, and staff expertise.}
}

\title{\LARGE \bf
VLM See, Robot Do: \\
Human Demo Video to Robot Action Plan via Vision Language Model
}

\maketitle
\thispagestyle{empty}
\pagestyle{empty}

\begin{abstract}
Vision Language Models (VLMs) have recently been adopted in robotics for their capability in common sense reasoning and generalizability. Existing work has applied VLMs to generate task and motion planning from natural language instructions and simulate training data for robot learning. In this work, we explore using VLM to interpret human demonstration videos and generate robot task planning. Our method integrates keyframe selection, visual perception, and VLM reasoning into a pipeline. We named it SeeDo because it enables the VLM to ``see'' human demonstrations and explain the corresponding plans to the robot for it to ``do''. To validate our approach, we collected a set of long-horizon human videos demonstrating pick-and-place tasks in three diverse categories and designed a set of metrics to comprehensively benchmark SeeDo against several baselines, including state-of-the-art video-input VLMs. The experiments demonstrate SeeDo's superior performance. We further deployed the generated task plans in both a simulation environment and on a real robot arm. 
The code, demos, prompts and data can be found at \href{https://ai4ce.github.io/SeeDo/}{ai4ce.github.io/SeeDo}.

\end{abstract}

\section{Introduction}
\label{sec:intro}
Large Vision Language Models (VLMs) have drawn significant interest in recent robotics and embodied AI research for their rich semantic and common-sense knowledge. Some research utilizes VLMs as an interface for parsing human language instructions to generate task plans~\cite{liang2023code,vemprala2024chatgpt,saycan2022arxiv,liu2024okrobot}. 
Some employ pretrained VLMs for further fine-tuning to learn the mapping from visual inputs and language instructions to actions~\cite{kim24openvla, black2024pi_0}, or leverage the general knowledge of VLMs to identify salient objects or key points~\cite{huang2023voxposer,huang2024rekep}.

\begin{figure}[t]
    \centering
    \includegraphics[width=\linewidth]{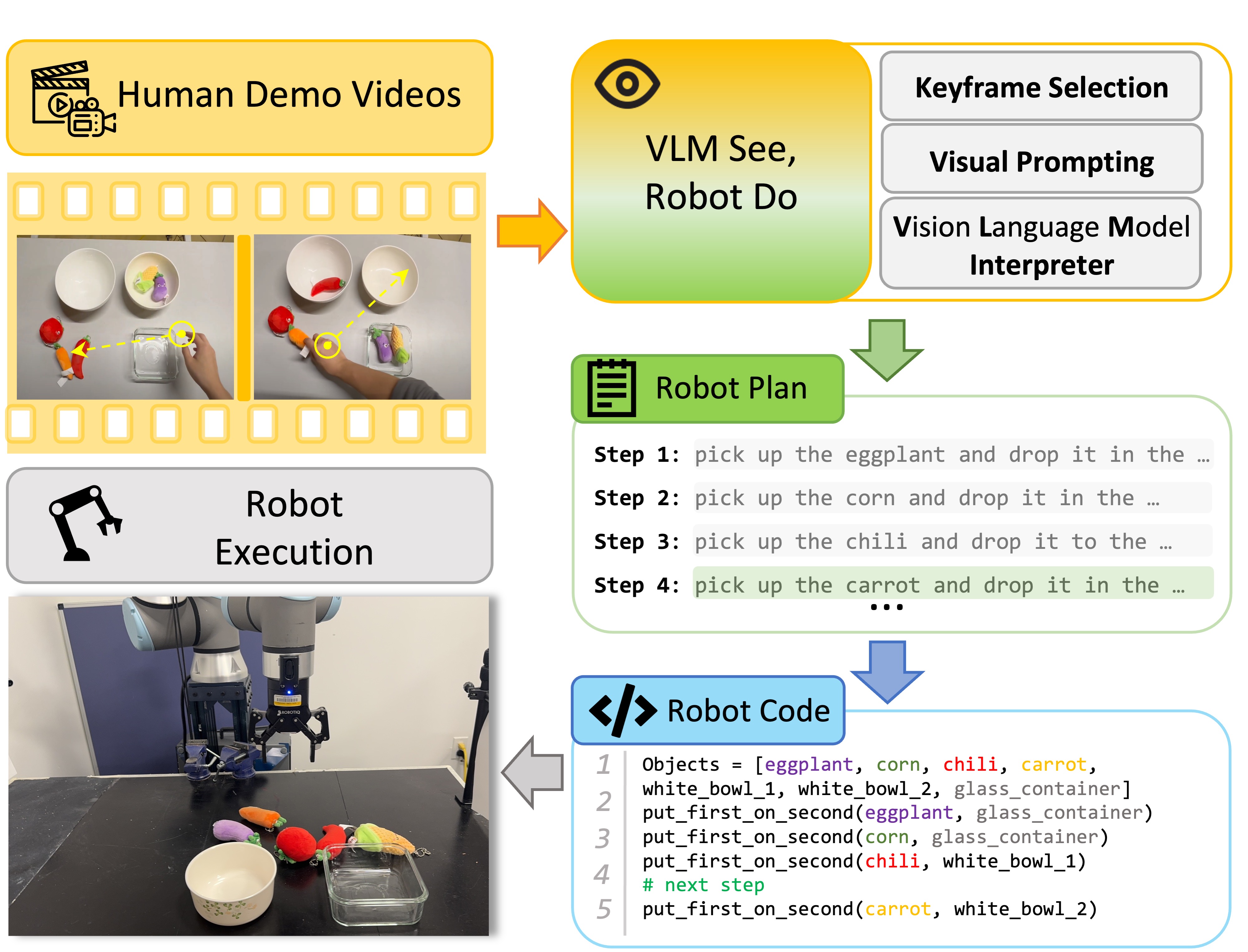}
    \vspace{-5mm}
    \caption{\textbf{VLM See, Robot Do.} We designed an agent framework centered around a large Vision Language Model to interpret long-horizon human demonstration videos into task plans in natural language, which are then executed in simulated and real-world robots via language model programs and action primitive functions.
    }
    \label{fig:teaser}
\vspace{-5mm}
\end{figure}

Although language instructions are effective in many scenarios, some tasks are inherently difficult to describe precisely using plain language. Additionally, as tasks become longer-horizon, providing instructions in language can become tedious and inefficient. In contrast, videos offer a more straightforward medium, making them particularly well-suited for long-horizon tasks that involve multiple steps or require understanding of temporal and spatial dependencies.
Moreover, video data is abundant on the internet and holds the promise to be leveraged in scaling robot learning~\cite{collaboration2023openx}. However, significant challenges remain in teaching robots to learn from human videos due to the substantial domain gap between robots and humans. While research in imitation learning tries to close the gap, long-horizon tasks often require collecting an increasing number of demonstrations. 
In contrast, humans can effectively learn new tasks by watching a few demonstration videos and intuitively breaking them down into sequential steps. Inspired by this capability, we propose \acronym, a modularized agent centered around a VLM. \acronym is designed to interpret long-horizon human demonstration videos into sub-task steps in natural language, which can then be executed by language model programs (LMPs)~\cite{liang2023code} and low-level action primitives.

Compared to directly feeding long-horizon human demonstration videos into imitation learning models, interpreting them with \acronym offers several advantages thanks to the capabilities of VLMs. First, VLMs' rich commonsense knowledge enables them to understand objects and their relationships, helping robots understand the task goals despite the embodiment gap. Second, their strong zero-shot generalization ability makes them more robust to unseen environments and objects during execution. The plans generated by VLMs remain valid even when object appearances, locations, or the surroundings differ between demonstration videos and real-world deployment.

Despite these advantages, we find VLMs struggle with processing a large number of video frames. They also lack the ability to accurately recognize objects and their spatial relations from the video, which is key information in robotics applications. To mitigate these limitations, \acronym integrates not only with a VLM interpreter module but also with a keyframe selection module and a visual prompting module, as illustrated in Figure~\ref{fig:teaser}.
The keyframe selection module identifies critical frames based on hand-speed heuristics, while the visual prompting module enhances the VLM’s ability in visual and spatial perception.
To evaluate \acronym, we curated a benchmark of \textit{long-horizon} pick-and-place tasks with human demonstration videos in three diverse categories: vegetable organization, garments organization, and wooden block stacking. All exhibit strong temporal and spatial dependencies.
We compare \acronym against baselines such as the state-of-the-art video VLMs~\cite{li2024llava, reid2024gemini}, and find that \acronym achieve the best performance. 

In summary, the contributions of this work are as follows:
\begin{itemize}
    \item We introduce \acronym, a VLM-based agent that integrates keyframe selection, visual prompting, and VLM interpreter modules to interpret long-horizon human demonstration videos and generate task plans for robots.
    \item We curate a benchmark of long-horizon human demo videos of pick-and-place tasks across three diverse categories and propose new evaluation metrics.
    \item We show that \acronym achieves superior results compared to state-of-the-art baselines that directly process videos or images as input. Furthermore, the generated task plans can be successfully deployed in both simulation and real-robot scenarios.
\end{itemize}

\section{Related Works}
\label{sec:related}
\textbf{VLMs for task planning.}
Large language models have shown amazing emergent abilities and generalizability to new tasks since the release of GPT-3~\cite{brown2020gpt3}, and have been used in generating task plans~\cite{huang2022language, guan2023leveraging, valmeekam2023planning, silver2024generalized}. To generate valid task planning, this body of work usually requires carefully curated instructions to prompt the VLMs~\cite{wei2022chain} and often relies on access to advanced close-sourced VLMs such as GPT-4~\cite{achiam2023gpt4} for good performance~\cite{silver2024generalized}. The task planning ability of VLMs has also been explored for robotics control~\cite{michal2024ecot, huang2022language, yoneda2024statler}.
One line of work generates robot executable codes as the medium of task plans~\cite{liang2023code, huang2022inner}. In these works, the VLMs take human language instructions and sometimes images as inputs, reason the instructions, and output the task plans as function-calling codes referred to as language model programs (LMPs), which call the wrapped API of robotics action primitives. 
\cite{du2023video} builds a video-language planning pipeline by using an embodied VLM~\cite{driess2023palme} to break long-horizon language instructions into steps, prompt a video model to generate video rollouts of the future, and assess current progress.
They rely on human language instructions to specify the task while \acronym explores directly interpreting long-horizon real-world human demonstration videos into the task specification.

\textbf{VLMs for robots.}
With the help of strong common sense understanding and rich semantic knowledge exhibited by the pretrained VLMs~\cite{bubeck2023sparks}, a line of literature~\cite{saycan2022arxiv, huang2023grounded, huang2022inner} has demonstrated that robots can take non-expert human language instructions more effectively than before~\cite{tellex2020robots}. Robotics researchers also go beyond using pretrained VLMs and leverage robotics data to train vision-language-action models (VLAs) catered for direct robotics control~\cite{driess2023palme, brohan2023rt2, kim24openvla}. Harnessing the rich common sense knowledge compressed in the large VLMs, useful data generation pipelines can be built to generate simulation data for training intelligent robots~\cite{yang2024unisim, wang2023robogen}. By leveraging the semantic knowledge of VLM, important objects or key points for execution can be selected and assigned affordance or constraint functions, thereby transforming the robot execution problem into an optimization problem over these functions.~\cite{huang2023voxposer,huang2024rekep}. 
In this work, we explore leveraging VLMs as an agent to the robots, interpreting human videos into task plans that can be further executed via LMPs.

\textbf{Robot learning from human videos.}
Demonstration videos provide a direct supervision signal for robot learning.
Many existing works leverage teleoperated robotics videos to train robot policies via imitation learning~\cite{collaboration2023openx, zhao2023act, fu2024mobile}.
For its massive amount compared to simulation and teleoperation, human video data has always held the promise of scaling up robot learning~\cite{nair2022r3m}. There has been a long interest in robot learning from human demonstrations from the early stage of robot learning~\cite{kaiser1996building, dillmann2004teaching} to the recent deep imitation learning~\cite{bahl2022human, bahl2023affordances, mendonca23swim, wang2023mimicplay}. 
Research in one-shot imitation learning~\cite{duan2017one, yu2018one,  bonardi2020learning, mandi2022towards, DBLP:journals/corr/abs-1909-04312, heppert2024ditto} aims to learn robot policies from a single demonstration, but they are limited in generalizability and confined to short-horizon tasks.
In general, having robots directly learn from human demonstration videos is still challenging due to the big domain gap between humans and robots. It also struggles with changes in the environment and object appearances between videos and deployments and often requires collecting many more demonstrations when the tasks become long-horizon~\cite{shafiullah2023dobbe}. 
To mitigate these challenges, \cite{wang2023mimicplay} proposes to break down the imitation learning process into training a latent planner to predict hand trajectory from human play data and training the planner-guided imitation policies on robotics data.
While sharing a similar motivation of decomposing planning and action learning, our work focuses on the planning part and takes a different approach. \acronym leverages pretrained VLMs to directly interpret human demonstration videos into text plans and the generated plans are processed into LMPs to call any action primitives whether they are trained-based, control-based, or pre-programmed.

\textbf{VLMs for video understanding.}
Recent VLMs have been trained to accommodate multiple modalities of inputs including videos~\cite{reid2024gemini, lin2024vila,zhang2024llavanextvideo,li2024llava} and can do video analysis tasks like question answering (QA) and video captioning. Video analysis~\cite{fu2024video} has also been included as part of the benchmark set to evaluate VLMs.
Our approach resembles a video QA or captioning setup but is grounded specifically in robotics scenarios. We also included several top-performing proprietary and open-source VLMs on the VideoMME~\cite{fu2024video} benchmark as our baselines.

\section{Method}
\label{sec:method}
\begin{figure*}[t]
    \centering
    \includegraphics[width=\linewidth]{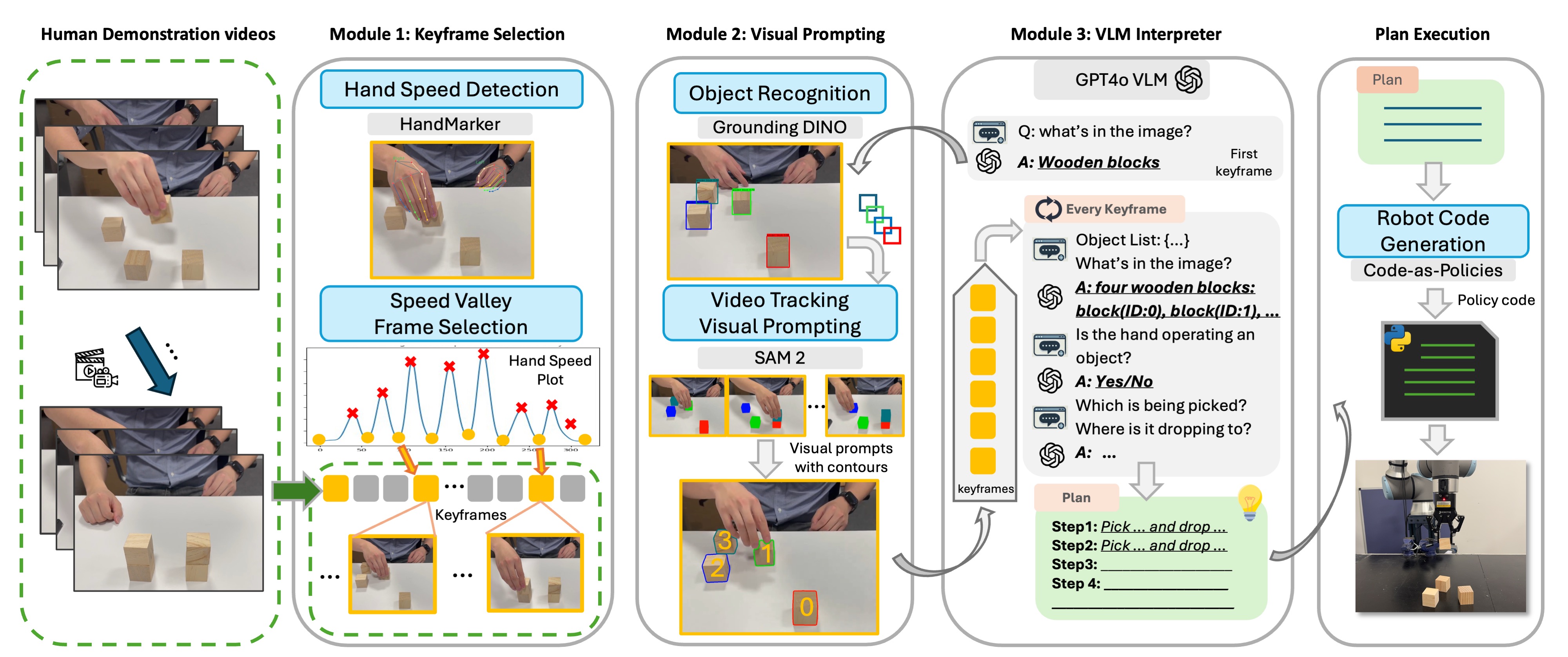}
    \vspace{-5mm}
    \caption{\textbf{The \acronym agent} consists of three modules. From left to right, a) The Keyframe Selection module detects the operating hand in the video and plots its speed. The speed valleys are identified as keyframes. b) The Visual Prompting module detects and tracks objects and then applies the tracking results as visual prompts to each keyframe. c) The VLM Interpreter module instructs the GPT-4o to interpret keyframes, identify objects and actions in each keyframe, and generate task plans from the demonstration video. d) Plan Execution. The generated task plans are processed by Code-as-Policies into language model programs (LMPs) and call the robot APIs for execution.}
    \label{fig:main}
\end{figure*}
In preliminary studies, we find that simply instructing pretrained VLMs with human demonstration videos yields poor results.
They constantly struggle with processing and retaining information from all frames, often confusing objects' temporal order and spatial relationships.
Therefore, we design an agent framework centered around the VLM to enhance its capabilities.
As a result, \acronym comprises three modules: \textit{Keyframe Selection module, Visual Prompting module, and VLM Interpreter module}. The full pipeline is illustrated in Figure~\ref{fig:main} and we detail each module below.

\textbf{Keyframe Selection.}
Context length becomes a major constraint when VLMs process long-horizon videos. Open-source VLMs often simply sample frames uniformly~\cite{li2024llava, lin2024vila}, but this approach is less effective for long-horizon demonstration videos, as frames showing important actions may be missed. To overcome the context limit yet still capture critical information, we adopt a heuristic approach that selects keyframes by detecting hand speed. Hand-object interactions are critical in demonstration videos~\cite{bahl2022human, bahl2023affordances} and we observe that hands typically move slower when picking or placing objects, providing a clue to locating keyframes~\cite{yanokura2020understanding}. 
Specifically, we use a lightweight model~\cite{lugaresi2019mediapipe} to detect hand keypoints, and plot the speed of the keypoints' center over time following~\cite{fang2024egopat3dv2}. 
The resulting plot is linearly interpolated and smoothed, producing a wave-like plot of hand speed. The frames corresponding to the troughs of the wave selected as keyframes.
Since hand detection is not always perfect and some troughs may be a result of interpolation,  we further filter out irrelevant keyframes. In the VLM planner module, we prompt the VLM to filter out the frames showing no hand-object interactions, which yields reasonably accurate results.

\textbf{Visual Prompting.}
Visual shortcomings of current VLMs have been reported in the literature~\cite{tong2024eyes}. We also find that VLMs often struggle to accurately determine object locations and their spatial relationships. They also frequently fail to consistently differentiate between visually similar objects over time, which is crucial for task planning from long-horizon human demonstration videos.
To alleviate these issues, we introduce a visual prompting module in \acronym that enhances the visual capabilities of the VLM. The module first instructs the VLM to identify objects in the frames and then use an open-vocabulary object detector~\cite{ren2024grounding} to extract object bounding boxes in the first frame. These bounding boxes are prompted to the Segment Anything Model (SAM2)~\cite{ravi2024sam2} for video tracking. The resulting tracking IDs and mask contours are rendered to the previously selected keyframes as visual prompts~\cite{bahng2022visualprompt}. These visual-prompted keyframes are subsequently given to the VLM Interpreter module.

\begin{figure*}[t]
    \centering
    \includegraphics[width=\linewidth]{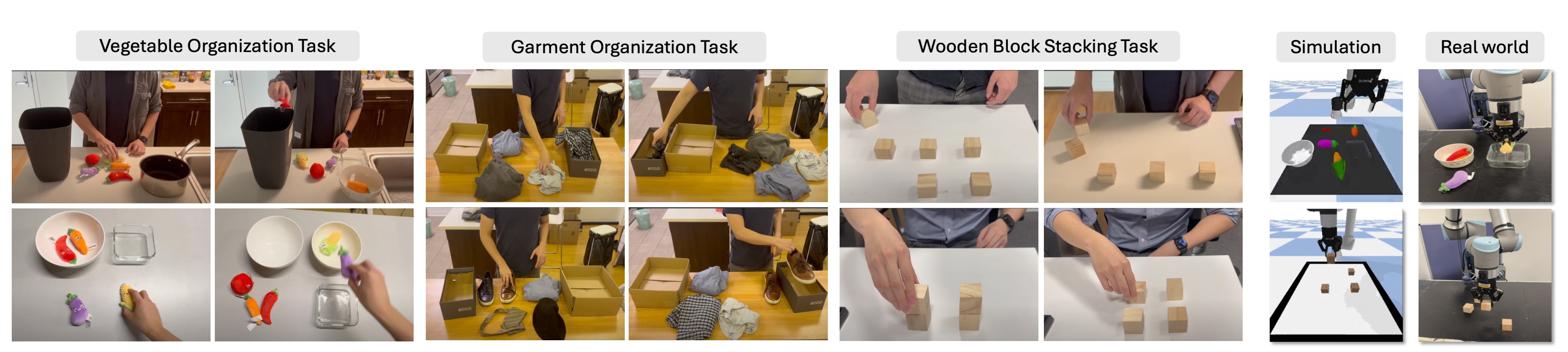}
    \vspace{-5mm}
    \caption{We collect long-horizon human demonstration videos across three diverse categories as our benchmark and carry out both simulation and real-world experiments. Tasks from left to right: vegetable organization, garment organization, and wooden block stacking.}
    \label{fig:data}
    \vspace{-5mm}
\end{figure*}

\textbf{VLM Interpreter.}
The VLM Interpreter module leverages chain-of-thought (CoT)~\cite{wei2022chain} to generate task planning steps for robot execution.
Any VLM can fit into the module, we choose GPT-4o for its popularity and widely recognized performance. We note that using closed-source models is not uncommon in the research community. Previous work in VLM for robotics~\cite{saycan2022arxiv, vemprala2024chatgpt, liang2023code, knowno2023, yoneda2024statler, dai2024think} has mostly adopted the GPT series or other close-sourced models such as the PALM~\cite{chowdhery2022palm, driess2023palme}.
A key design is the incorporation of visual prompts. Previous studies have shown that providing VLMs with images containing prompted visual cues is effective for manipulation~\cite{liu2024moka} and navigation tasks~\cite{sathyamoorthy2024convoi}. In \acronym, identifying objects and understanding their spatial relationships is crucial to interpret long-horizon human demonstration videos, and this is greatly enhanced by the visual prompts derived from video tracking in the Visual Prompting module.
Specifically, we maintain an object list obtained from the Visual Prompting module in the text prompt. The mask contours and tracking IDs are used as visual prompts in the keyframes to aid object identification.
By using contours instead of full masks we ensure the objects of interest are highlighted without obstructing their appearance. 
The center coordinates of the masks, combined with the corresponding tracking IDs, are then appended to the textual prompts to imply the VLM of the objects' spatial relationships. We find this design particularly helpful when the human demonstration videos involve visually similar objects, such as when playing with uncolored wooden blocks.
Below is an example CoT prompt in \acronym. The complete prompt will be made available along with the code.

\textbf{Prompt1:} Example of CoT Prompt Design

\vspace*{-5pt}  
\begin{coloredlisting}
# CoT step1: get object list
(*@\colorbox{yellow!25}{\parbox{0.97\linewidth}{How many objects are there on the desk? \\ What are these objects?}}@*)
(*@\colorbox{ForestGreen!25}{\parbox{0.97\linewidth}{ - num:5, objects:[red chili, orange carrot, white bowl]}}@*)
# CoT step2: choose object picked
(*@\colorbox{yellow!25}{\parbox{0.98\linewidth}{Which object is being picked among [red chili, orange carrot, white bowl]?}}@*)
(*@\colorbox{ForestGreen!25}{\parbox{0.98\linewidth}{ - object picked: red chili}}@*)
# CoT step3: choose reference object
(*@\colorbox{yellow!25}{\parbox{0.98\linewidth}{Which object among [red chili, orange carrot, white bowl] do you choose as reference object?}}@*)
(*@\colorbox{ForestGreen!25}{\parbox{0.98\linewidth}{ - reference object: white bowl}}@*)
# CoT step4: choose where to drop
(*@\colorbox{yellow!25}{\parbox{0.98\linewidth}{In which direction do you drop red chili to the white bowl?}}@*)
(*@\colorbox{ForestGreen!25}{\parbox{0.98\linewidth}{ - Drop red chili in the white bowl}}@*)
\end{coloredlisting}

\textbf{Plan Execution.}
The \acronym-generated task plans can be seamlessly processed step by step by any robot action model that can take text input. Specifically, following the approaches in~\cite{liang2023code} and~\cite{yoneda2024statler}, we use Language Model Programs (LMPs) to implement the task plans on a UR10e robot arm in both Pybullet simulations~\cite{coumans2016pybullet} and real-world deployment. Action primitives are predefined as functions and can be called by the LMPs. In real-world experiment, we follow \cite{liang2023code, yoneda2024statler} and first use a segmentation model to segment all objects of interest in the RGB images, then query the depth image with the same coordinates to acquire 3D information of the objects.

\section{Experiments}
\label{sec:exp}

\begin{table*}[t]
\vspace{2mm}
  \centering
  \caption{Model Success Rate Across Different Tasks}
  \label{tab:model_performance}
  \setlength{\tabcolsep}{13pt}
  \begin{tabular}{@{}lcccccccccc@{}}
    \toprule
     \multirow{2}{*}{\textbf{Model}} & \multicolumn{3}{c}{\textbf{Vegetable Organization}} & \multicolumn{3}{c}{\textbf{Garment Organization}} & \multicolumn{3}{c}{\textbf{Wooden Block Stacking}} \\ \cmidrule(lr){2-4} \cmidrule(lr){5-7} \cmidrule(lr){8-10}
    & \textbf{TSR} & \textbf{FSR} & \textbf{SSR} & \textbf{TSR} & \textbf{FSR} & \textbf{SSR} & \textbf{TSR} & \textbf{FSR} & \textbf{SSR} \\
    \midrule
    \textbf{LLaVA-OneVision~\cite{li2024llava}} & 0.00 & 0.00 & 2.75 & 0.00 & 0.00 & 31.61 & 0.00 & 0.00 & 2.56 \\
    \textbf{LLaVA-NeXT-Video-7B~\cite{zhang2024llavanextvideo}} & 0.00 & 0.00 & 9.03 & 0.00 & 0.00 & 4.00 & 0.00 & 0.00 & 3.85 \\
    \textbf{VILA1.5-8B~\cite{lin2024vila}} & 5.26 & 5.26 & 14.38 & 0.00 & 0.00 & 0.90 & 0.00 & 0.00 & 3.41 \\
    \textbf{Gemini 1.5 Pro~\cite{reid2024gemini}} & 39.47 & 39.47 & 70.00 & 16.67 & 16.67 & 57.22 & 0.00 & 0.00 & 13.80 \\
    \textbf{GPT-4o~\cite{achiam2023gpt4} Init + Final} & 39.47 & 39.47 & 54.43 & 13.33 & \textbf{30.00} & 31.61 & 10.26 & 15.38 & 32.69 \\
    \textbf{\acronym} & \textbf{60.53} & \textbf{60.53} & \textbf{80.40} & \textbf{26.67} & 26.67 & \textbf{66.50} & \textbf{21.62} & \textbf{21.62} & \textbf{52.48} \\
    \bottomrule
  \end{tabular}
\end{table*}

\begin{figure}[htbp]
    \centering
    \includegraphics[width=\linewidth]{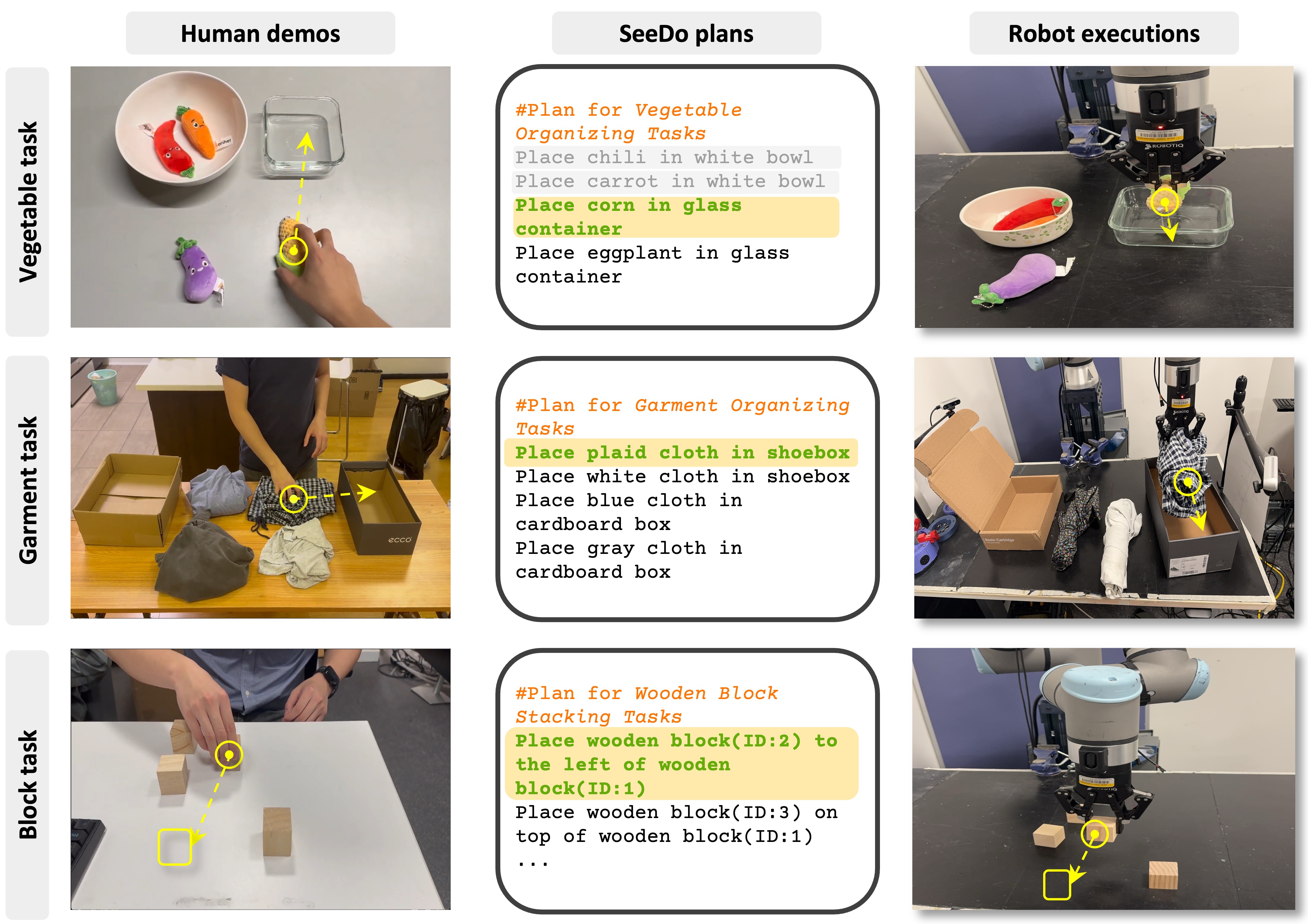}
    \caption{\textbf{Results visualization on all three tasks.}
    }
    \label{fig:real world experiments}
\end{figure}

We curated long-horizon human demonstration videos across three diverse pick-and-place tasks and designed a set of metrics for comprehensive evaluation. In addition to comparing \acronym to baselines across all three tasks. We also present ablation studies to assess the impact of separate modules. Finally, we analyzed and discussed the types of errors that occurred with \acronym and the baselines. Qualitative results are shown in Figure~\ref{fig:real world experiments}.

\subsection{Task Design}
We focus on long-horizon daily and construction tasks that decompose naturally into a series of pick-and-place sub-tasks. These tasks represent some common robotics scenarios that feature a clear temporal sequence and dynamic interactions that cannot be adequately captured with still images or brief descriptions. Their reliance on sequential dependencies makes it nearly impossible to specify or replicate the tasks using static visuals or a few words, underscoring the need for full demonstration videos.

As illustrated in Figure~\ref{fig:data}, we curate a set of human demonstration videos covering three diverse categories as the evaluation tasks:

\textbf{Vegetable Organization Task} contains demonstration videos showing humans picking up and dropping different kinds of vegetables into several different containers. There are 6 different vegetables and 4 different containers. The containers include a ceramic bowl, a glass container, a trash bin, and a small pot to best mimic real-life kitchen scenarios. For the vegetables, we use plush toys instead of real ones as in~\cite{michal2024ecot} to avoid potential damage in real experiments. In the simulation deployment, we use Pybullet~\cite{coumans2016pybullet} and collected online free \texttt{.obj} models of some objects and utilize a publicly available text-to-3D generation model~\cite{lumaAI_genie_2023} to generate the others. In total, there are 38 demonstrations.   

\textbf{Garment Organization Task} contains demonstrations of a human organizing their garments into separate boxes. Garments are visually distinct from the vegetables and serve as a complement to the vegetable task in the daily scenario. To make the task interesting and challenging, we chose garments of various types including shirts, shoes, ties, and an umbrella. In total we collected 30 demonstrations.

\textbf{Wooden Block Stacking Task}. We also collected demonstrations of a human playing with wooden blocks to simulate a block-building game-play scenario or a miniature construction site setup. The key feature of this category is that the visual appearance of the objects is highly similar, which requires relatively precise understanding of spatial relations, creating a well-known challenging case for the current vision language models. We show that \acronym can overcome this challenge with the help of the visual prompts from the Visual Prompting module. 
In total, this task contains 39 demonstrations.

Our method operates purely on relative spatial relationships (e.g., left, right, above, below) extracted from the demonstration, without relying on fixed camera viewpoints; thus, the execution replicates the same relative positions regardless of whether the input is from a first-person or third-person perspective.

\subsection{Evaluation Metrics}\label{sec:metric}
Pick-and-place tasks are fundamental building blocks of object manipulation. 
Conventional evaluation metric reports success rate (SR) of each task which could only reflect the completion at the final state of operation.
We are particularly interested in how well the robot can follow the demonstration videos step by step. Therefore, we propose our metrics.

Specifically, a long-horizon pick-and-place task can be decomposed into a series of single pick-and-place sub-task steps arranged sequentially over time. Each pick-and-place action establishes a relative spatial relation pair between two objects. The final state after the completion of the entire long-horizon task is thus a set of these relational pairs. This provided a measurement of the alignment between the sequence of pick-drop steps generated by a model and those demonstrated in the human videos. Based on this, we propose three metrics: Task Success Rate (TSR), Final-state Success Rate (FSR), and Step Success Rate (SSR), to evaluate the completeness of the generated task plan. 
\begin{itemize}
    \item \textbf{TSR} strictly evaluates if a generated plan exactly aligns with the video. To achieve success (TSR=1), each step in the plan must match the demo's action sequence in both content and temporal order.
    \item \textbf{FSR} is equivalent to the conventional SR in that it relaxes from TSR and marks a success as long as the final state of objects matches the result of the demonstration, regardless of the temporal order of execution.
    \item \textbf{SSR} evaluates the partial completeness. It aligns the pick-drop steps from the generated plan to the demo video in temporal order and computes the ratio of the number of aligned steps over the total number of groundtruth steps suggested in the demo. In other words, it implies how far the plan goes before a mistake occurs.
\end{itemize}
Additionally, we identify three types of errors from the failure cases to analyze and provide insights on the strengths and weaknesses of various models on our tasks:
\begin{itemize}
    \item \textbf{Vision Error} occurs when a model fails to recognize or effectively distinguish between different objects, reflecting the model's capability on visual recognition.
    \item \textbf{Spatial Error} occurs when the objects are correctly identified and distinguished, but there is an error in understanding the spatial relations between them, reflecting the spatial understanding capability.
    \item \textbf{Temporal Error} occurs when the number of actions in the output differs from that in the human demonstration, or when the temporal order of actions is incorrect, reflecting issues with a model's capability in video understanding and temporal reasoning.
\end{itemize}

\subsection{Baselines}
\textbf{Video understanding VLMs}. We compare \acronym with VLMs that are capable of direct video understanding. This includes the proprietary Gemini 1.5 Pro, a leading model for video understanding~\cite{fu2024video}, as well as several open-source VLMs that also exhibit strong video understanding abilities, including LLaVA-OneVision~\cite{li2024llava}, LLaVA-NeXT-Video-7B-DPO and VILA~\cite{lin2024vila}. For consistency, we use the same core prompts as \acronym across all baselines. Gemini 1.5 Pro natively supports video input, while for the open-source models, we follow their official implementations, uniformly sampling 16 frames per video as input. For the open-source baselines, prompts are slightly adjusted to optimize each model’s performance, and the full prompt details will be released with our code.

\textbf{GPT-4o with alternative frame sampling strategies}. 
Since \acronym utilizes GPT-4o as its VLM, we further test three variants of GPT-4o using different frame sampling strategies while keeping the same prompts: 
\begin{itemize}
\item \textit{GPT-4o Init+Final}: Uses only the first and last frames of each video to assess whether providing video input is necessary or if start- and end-state images alone suffice
\item \textit{GPT-4o Unif.}: Following the method in ~\cite{Wake_2024}, uniformly samples 16 frames without hand detection,  to ablate the Keyframe Selection module.
\item \textit{GPT-4o without Visual Prompting}: Evaluates the impact of removing the Visual Prompting module.
\end{itemize}
These two ablation baselines are reported respectively in Table~\ref{tab:sampling strategy results} and Table~\ref{tab:ablation_visual_perception_success_rate}.

\subsection{Results}
\textbf{Main results.} Table~\ref{tab:model_performance} presents the overall results of \acronym and baselines on all three tasks. \acronym~outperforms all closed-source and open-source video VLM baselines across TSR, FSR, and SSR. Since the demonstration videos are all long-horizon, both TSR and FSR require a precise understanding of the entire pick-and-place task from the full-length video. This imposes high accuracy demands on all three modules of \acronym. Consequently, we observe that when temporal order is incorrectly comprehended, the final state is also wrong.
An exception is the GPT-4o Init+Final experiment, where its FSR on the garment task is slightly higher. This shows GPT-4o's strong common sense. However, its overall TSR and SSR are still poor, indicating a weakness in temporal reasoning.
Meanwhile, we observed that even in cases where TSR and FSR fail, \acronym~is often able to successfully interpret most of the task steps in the correct temporal order. As a result, \acronym's SSR accuracy exceeds 70 percent for the two daily tasks and 50 percent for the challenging block stacking tasks.

\textbf{Ablation on keyframe selection.} To ablate on our hand-detection-based Keyframe Selection module, in Table~\ref{tab:sampling strategy results}  we present the experiment using uniform sampling for keyframe extraction. We found that, despite using the same VLM model, GPT-4o, its output is almost always incorrect.
We find that it is difficult for this method to ensure that the keyframes indicating crucial actions are sampled, which negatively impacts GPT-4o's understanding.
Moreover, the context limit is often exceeded when inputting all sampled frames at once.
In contrast, hand detection for keyframe selection can effectively extract the key information of the demonstrations and yield superior performance.

\begin{table}[t]
\vspace{2mm}
  \centering
  \caption{Different sampling. (T/F/S) stands for TSR, FSR, and SSR. I+F stands for initial+final frames. Unif. stands for uniform frame samplings. 
  }
  \label{tab:sampling strategy results}
  \setlength{\tabcolsep}{2.0pt}
  \begin{tabular}{@{}lcccccccccc@{}}
    \toprule
    \multirow{2}{*}{\textbf{Model}} & \multicolumn{3}{c}{\textbf{Vegetable}} & \multicolumn{3}{c}{\textbf{Garment}} & \multicolumn{3}{c}{\textbf{Block}} \\ 
    \cmidrule(lr){2-4} \cmidrule(lr){5-7} \cmidrule(lr){8-10}
    & \textbf{T} & \textbf{F} & \textbf{S} & \textbf{T} & \textbf{F} & \textbf{S} & \textbf{T} & \textbf{F} & \textbf{S} \\
    \midrule
    \textbf{GPT-4o I+F} & 39.47 & 39.47 & 54.43 & 13.33 & \textbf{30.00} & 31.61 & 10.26 & 15.38 & 32.69 \\
    \textbf{\acronym\ Unif.} & 0.00 & 0.00 & 1.32 & 0.00 & 0.00 & 0.67 & 0.00 & 0.00 & 0.00 \\
    \textbf{\acronym} & \textbf{60.53} & \textbf{60.53} & \textbf{80.40} & \textbf{26.67} & 26.67 & \textbf{66.50} & \textbf{21.62} & \textbf{21.62} & \textbf{52.48} \\
    \bottomrule
  \end{tabular}
\end{table}

\textbf{Ablation on the visual prompting for Spatial Understanding.} For pick-and-place task, describing the relative spatial positions between objects is crucial. Our wooden block stacking tasks in essence require more precise spatial understanding. To explore the impact of visual prompts on the understanding capability for complex relative spatial relationships, we experimented \acronym~without visual prompt (SeeDo w/o V.P.) as the baseline to compare with the \acronym with visual prompting, which uses mask contours, tracking IDs, and the corresponding coordinates of contour centers as prompts. Experiment results on the wooden block task are shown in Table~\ref{tab:ablation_visual_perception_success_rate}. The contrast in the percentage of vision and spatial errors suggests that visual prompting significantly enhances spatial understanding capabilities.

\begin{table}[t]
  \centering
  \caption{Ablation on visual prompting (V.P.) on Wooden Block Stacking Task.}
  \label{tab:ablation_visual_perception_success_rate}
  \setlength{\tabcolsep}{3.7pt}
  \begin{tabular}{@{}lcccccc@{}}
    \toprule
    \multirow{2}{*}{\textbf{Model}} & \multicolumn{3}{c}{\textbf{Success Rate}} & \multicolumn{3}{c}{\textbf{Failure Reason}} \\ 
    \cmidrule(lr){2-4} \cmidrule(lr){5-7}
    & \textbf{TSR}$\uparrow$ & \textbf{FSR}$\uparrow$ & \textbf{SSR}$\uparrow$ & \textbf{Vision}$\downarrow$ & \textbf{Spatial}$\downarrow$ & \textbf{Temporal}$\downarrow$ \\
    \midrule
    \textbf{\acronym w/o V.P.} & 0.00 & 0.00 & 41.67 & 42.86 & 87.5 & 28.57 \\
    \textbf{\acronym w/ V.P.} & \textbf{21.62} & \textbf{21.62} & \textbf{52.48} & \textbf{20.51} & \textbf{64.10} & \textbf{17.95} \\
    \bottomrule
  \end{tabular}
\end{table}

\begin{figure}[t]
    \centering
    \includegraphics[width=\linewidth]{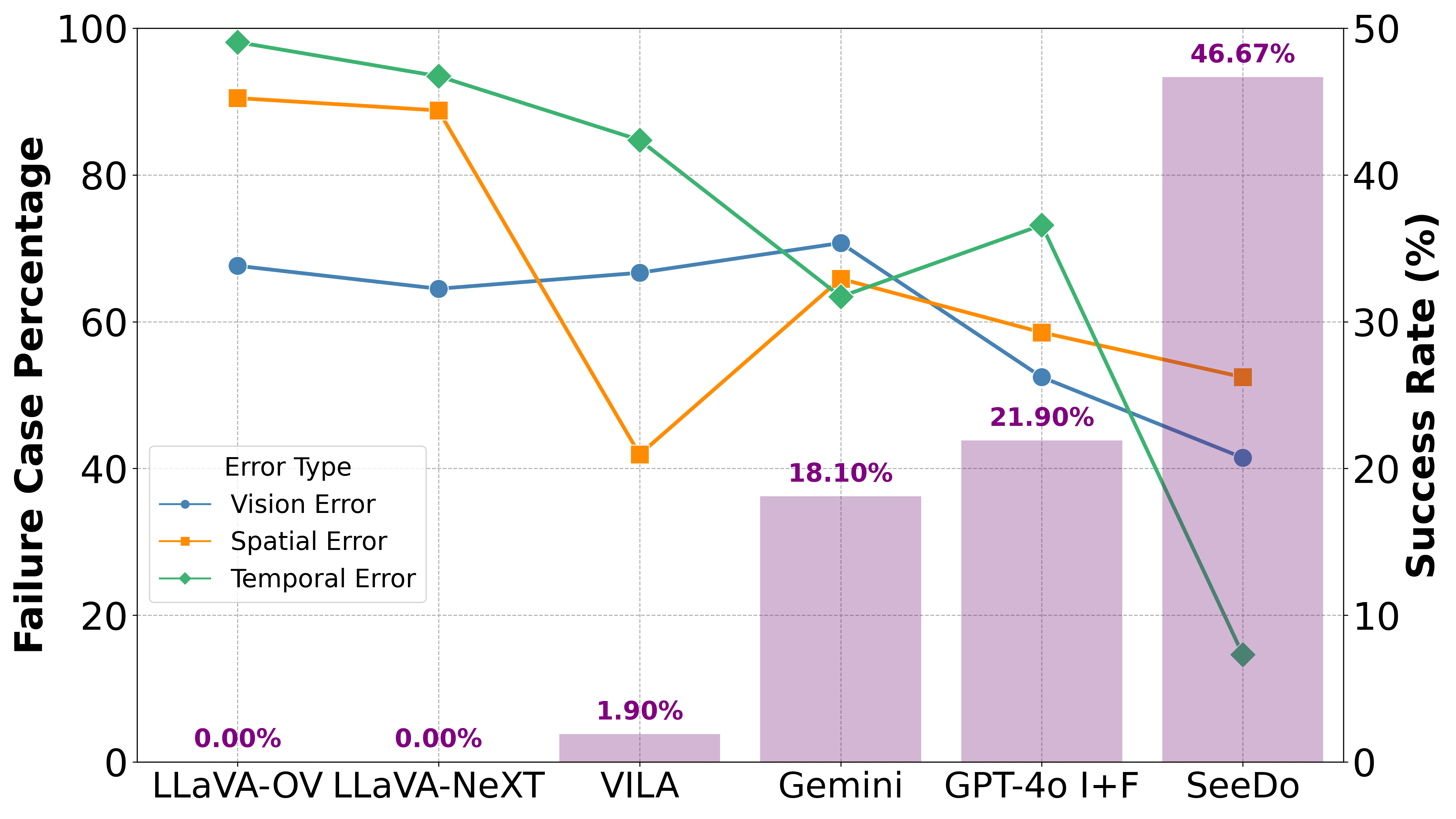}
    \vspace{-7mm}
    \caption{\textbf{Error type percentages} of all the failure cases of all the methods. Note that error types are not exclusive. The barplot of the total success rates on all tasks is also presented. LLaVA-OV represents the LLaVA-OneVision model.
    }
    \label{fig:error_type_comparison}
\end{figure}

\textbf{Failure Case Analysis.}
To better understand the performance, we categorized failure cases based on the three error types discussed in Sec.~\ref{sec:metric} and calculated their frequency across all demonstrations. As shown in Figure~\ref{fig:error_type_comparison}, \acronym consistently has lower error rates across all categories compared to other models, particularly excelling in a notably lower temporal error rate. However, spatial errors remain the main source of \acronym’s failures.
This could be largely attributed to the limited spatial intelligence of current VLMs and the imperfect tracking in the visual prompting module. 
Occasionally, mismatches occur between the two modules, where tracking associates an object with a text description that the VLM refers to as another. This indicates room for future improvement.
It is worth noting that the error types are not mutually exclusive and there is a coupling effect. For instance, VILA has a high temporal error rate due to frequently generating erroneous plans with repetitive steps or omitting too many steps. These failure cases are marked as temporal errors, leading to a lower relative incidence of spatial errors. This should not be interpreted as evidence of a strong spatial understanding in VILA, but rather as a consequence of its frequent temporal errors.

\subsection{Real-world Experiment Setup}
Real-world experiments have demonstrated that SeeDo can manipulate objects in the physical world using an appropriate LMP.
The experiment is conducted on a Universal Robots' UR10e cobot attached with a Robotiq 2F-85 gripper. We use an Intel RealSense 455 stereo camera to acquire depth information. The camera is mounted on a tripod standing across the robot and hand-eye calibrated in an eye-to-hand setup. The gripper is programmed to move horizontally above the object and lower in the z-axis to grasp the object. 

Because the action primitives are hard-coded, we find that sometimes inevitably result in some failure cases due to lack of adaptability. For example, in the wood block stacking example, the dropping distance is often too high and would cause the wood block to bounce upon contact, causing task failure in the real world. Such failure cases can be improved in the future by fine-tuning the robot's action primitives or substituting it with more adaptive and generalizable robot policies. In its current version, SeeDo's evaluation only measures the accuracy of the action plan and does not take into account the success rate of actual execution.

\section{Conclusion and Limitations}
\label{sec:conclusion}
This paper tackles the challenge of extracting robot task plans directly from human demonstration videos using vision language models (VLMs).
We introduce a pipeline, \acronym, that significantly improves temporal understanding, spatial relationship reasoning, and object differentiation, particularly in cases where objects have similar appearances, outperforming existing video VLMs.
Through comprehensive evaluation, \acronym demonstrates state-of-the-art performance on long-horizon tasks of a series of pick-and-place actions in diverse categories. 
However, \acronym still faces many limitations, and we discuss several below.

\textbf{Limited action space.} The current experiments are limited to pick-and-place actions. Extending \acronym to action spaces with more complex behavioral logic or a wider variety of behaviors is our next step.

\textbf{Limited spatial intelligence.} While the visual perception module significantly improves \acronym's ability to differentiate left and right spatial relations, it still makes mistakes in tasks requiring more precise spatial reasoning. We call for future VLMs with stronger spatial intelligence.

\textbf{Under-defined spatial positioning.} Describing the spatial positions of objects is inherently complex. In this work, \acronym only describes the relative position as a limited set of high-level relative spatial relation pairs, which makes it less competent for tasks requiring precise manipulation. In future work, we will explore extracting more precise spatial positioning from the demonstration videos.


\newpage
{
\small
\bibliographystyle{IEEEtran}
\balance
\bibliography{IEEEabrv}
}

\newpage
\section*{Appendix}
\label{sec:appendix}
\subsection{Prompt Engineering}
In SeeDo, the Vision Language Model (VLM) is primarily utilized in two key areas: the first is within the Visual Perception Module, where it extracts an object list from the environment and uses it as a prompt for the GroundingDINO to perform object detection; the second is in VLM Reasoning Module, where it generates the corresponding robot action plan based on the keyframes.
We chose \textit{gpt-4o-2024-08-06} as it was the VLM with state-of-the-art performance by the date of this paper.
\subsubsection{Visual Perception Prompts}
The following prompt is a prompt used in the Visual Perception Module to obtain the object list from the environment. In cases where objects with similar visual appearances are difficult to distinguish through language (e.g., multiple wooden blocks), the VLM is prompted to repeat such objects in the output according to their occurrence. For example, if there are three wooden blocks, the VLM is instructed to output "wooden block, wooden block, wooden block."
\begin{lstlisting}[language=Python, caption=Visual Perception Prompt, label=lst:visual_perception_prompt]
## Prompt VLM to obtain object list
[system prompt]
- You are a visual object detector. Your task is to count and identify the objects in the provided image that are on the desk. Focus on objects classified as grasped_objects and containers.
- Do not include hand or gripper in your answer.
[user prompt]
- There are two kinds of objects, grasped_objects and containers in the environment. We only care about objects on the desk.
- You must strictly follow the rules below: Even if there are multiple objects that appear identical, you must repeat their names in your answer according to their quantity. For example, if there are three wooden blocks, you must mention 'wooden block' three times in your answer.
- Be careful and accurate with the number. Do not miss or add additional object in your answer.
- Based on the input picture, answer:
1. How many objects are there in the environment?
2. What are these objects?
- You should respond in the format of the following example:
- Number: 4
- Objects: red pepper, red tomato, white bowl, white bowl
\end{lstlisting}

\subsubsection{VLM Reasoning Prompts}
VLMs tend to overlook the prompt content or fail to fully follow prompt instructions when dealing with longer contexts, which fundamentally poses challenges for understanding long-horizon human demonstration videos. At the same time, as the number of pick-and-place actions increases along with the length of the video, the number of keyframes in the keyframe list also increases. Through experimentation, we found that inputting all keyframes representing long-horizon critical actions into the VLM as a single prompt often exceeds the token limit of the ChatGPT API. Even for the rare, shorter keyframe lists that fall within the token limit, the VLM has been shown to be incapable of producing accurate results. To address this, we adopted a Chain of Thought (CoT) design approach in the prompt section of the VLM.

For a long-horizon pick-and-place task, the filtered keyframe sequence typically includes three types of keyframes: those showing the pick action, those showing the place action, and some mistakenly selected keyframes. The valid information lies in the pick and place keyframes. After eliminating the misselected keyframes, the remaining pick and place keyframes can be paired in sequence, with each pick keyframe followed by its corresponding place keyframe. Our Chain of Thought (CoT) design is based on this approach.
In the CoT process, the first step is to filter the keyframes, removing invalid frames where the hand is not interacting with any object. For valid frames, we first prompt the VLM to understand the pick frame, identifying the object picked in the pick-and-place subtask. Next, we pass both the object picked and the place frame to the VLM, instructing it to select the appropriate reference object. Once the reference object is determined, the VLM is then tasked with outputting the spatial relationship between the object picked and the reference object in the place step.
\begin{lstlisting}[language=Python, caption=Filter Invalid Keyframe, label=lst:filter_invalid_keyframe]
## Prompt VLM to filter invalid keyframes
[system prompt]
- You are an operations inspector. You need to check whether the hand in operation is holding an object. The objects have been outlined with contours of different colors and labeled with indexes for easier distinction.
[user prompt]
- This is a picture from a pick-and-drop task. Please determine if the hand is manipulating an object.
- Respond with 'Hand is manipulating an object' or 'Hand is not manipulating an object'.
\end{lstlisting}
In Prompt~\ref{lst:filter_invalid_keyframe}, if the respond is 'Hand is not manipulating an object', then this key frame is marked as invalid and ignored.
\begin{lstlisting}[language=Python, caption=Identify Object Picked, label=lst:idenfity_object_picked]
## Prompt VLM to obtain object picked
[system prompt]
- You are an operation inspector. You need to check which object is being picked in a pick-and-drop task. Some of the objects have been outlined with contours of different colors and labeled with indexes for easier distinction.
- The contour and index is only used to help. Due to limitation of vision models, the contours and index labels might not cover every objects in the environment. If you notice any unannotated objects in the demo or in the object list, make sure you name it and handle them properly.
[user prompt]
- This is a picture describing the pick state of a pick-and-drop task. The objects in the environment are {obj_list}. One of the objects is being picked by a human hand or robot gripper now. The objects have been outlined with contours of different colors and labeled with indexes for easier distinction.
- Based on the input picture and object list, answer:
1. Which object is being picked
- You should respond in the format of the following example:
- Object Picked: red block
\end{lstlisting}
The \{obj\_list\} in Prompt~\ref{lst:idenfity_object_picked} are extracted from the VLM response to Prompt~\ref{lst:visual_perception_prompt}. The \{object\_picked\} in Prompt~\ref{lst:identify_reference_object},~\ref{lst:reason_spatial_relationship} are extracted from the VLM response to Prompt~\ref{lst:idenfity_object_picked},~\ref{lst:identify_reference_object}.
\begin{lstlisting}[language=Python, caption=Identify Reference Object, label=lst:identify_reference_object]
## Prompt VLM to identify the reference object
[system prompt]
- You are an operation inspector. You need to find the reference object for the placement location of the picked object in the pick-and-place process. Notice that the reference object can vary based on the task. If this is a storage task, the reference object should be the container into which the items are stored. If this is a stacking task, the reference object should be the object that best expresses the orientation of the arrangement.
[user prompt]
- This is a picture describing the drop state of a pick-and-place task. The objects in the environment are {obj_list}. {object_picked} is being dropped by a human hand or robot gripper now.
- Based on the input picture and object list, answer:
1. Which object in the rest of object list do you choose as a reference object to {object_picked}
- You should respond in the format of the following example without any additional information or reason steps:
- Reference Object: red block
\end{lstlisting}
Once the object picked and reference object are selected, the VLM is then prompted to reason about the spatial relationship between these two objects. For our tasks, we define six different spatial relationships: in, on top of, at the back of, in front of, to the left, to the right.
In our task, these spatial relationships are defined to be \textbf{mutually exclusive}. The object picked and the reference object can exhibit multiple relative positional relationships across different dimensions in three-dimensional space. The VLM is tasked with selecting the most dominant relationship from six predefined relative spatial relationships based on its understanding and judgment.
\begin{lstlisting}[language=Python, caption=Reason Spatial Relationship, label=lst:reason_spatial_relationship]
## Prompt VLM to reason about the spatial relationship between object picked and reference object
[system prompt]
- You are a VLMTutor. You will describe the drop state of a pick-and-drop task from a demo picture. You must pay specific attention to the spatial relationship between picked object and reference object in the picture and be correct and accurate with directions.
[user prompt]
- This is a picture describing the drop state of a pick-and-drop task. The objects in the environment are object list: {obj_list}. {object_picked} is said to be being dropped by a human hand or robot gripper now.
- However, the object being dropped might be wrong due to bad visual prompt. If you feel that object being picked is not {object_picked} but some other object, red chili is said to be the object picked but you feel it is an orange carrot, you MUST modify it and change the name.
- The object picked is being dropped somewhere near {object_reference}. Based on the input picture, object list answer:
- Drop object picked to which relative position to the {object_reference}? You need to mention the name of objects in your answer.
- There are totally six kinds of relative position, and the direction means the visual direction of the picture. You must choose one relative position.
1. In ((object picked is contained in the reference object)
2. On top of (object picked is stacked on the reference object, reference object supports object picked)
3. At the back of (in demo it means object picked is positioned farther to the viewer relative to the reference object)
4. In front of (in demo it means object picked is positioned closer to the viewer or relative to the reference object)
5. to the left
6. to the right
- You should respond in the format of the following example without any additional information or reason steps, be sure to mention the object picked and reference object.
- Drop yellow corn to the left of the red chili.
- Drop wooden block (ID:1) to the right of the wooden block (ID:0)
\end{lstlisting}
\subsection{Evaluation Metrics}

We employ the following evaluation metrics to assess the performance of the model:

\textbf{Task Success Rate (TSR)}
The task success rate measures whether the predicted steps exactly follow the steps demonstrated in the video. The computation is defined as:

\begin{algorithm}[H]
\caption{TSR: Task Success Rate}
\begin{algorithmic}[1]
\STATE \textbf{Input:} Predicted steps $\mathcal{P}$, Ground truth steps $\mathcal{G}$
\STATE \textbf{Output:} TSR value
\IF{$\mathcal{P}$ exactly matches $\mathcal{G}$}
    \STATE \textbf{return} 1
\ELSE
    \STATE \textbf{return} 0
\ENDIF
\end{algorithmic}
\end{algorithm}

\textbf{Final State Success Rate (FSR)}
The final state success rate evaluates the final states of the objects specified as their relative spatial relations. Specifically, it checks if the predicted plan results in the same final states as the demonstration's. The computation is defined as:

\begin{algorithm}[H]
\caption{FSR: Final State Success Rate}
\begin{algorithmic}[1]
\STATE \textbf{Input:} Final predicted state $\mathcal{S}_{\text{final}}$, Final ground truth state $\mathcal{S}_{\text{final}}$
\STATE \textbf{Output:} FSR value
\IF{$\mathcal{S}_{\text{final}}$ exactly matches $\mathcal{S}_{\text{final}}$}
    \STATE \textbf{return} 1
\ELSE
    \STATE \textbf{return} 0
\ENDIF
\end{algorithmic}
\end{algorithm}

\textbf{Step Success Rate (SSR)}
The step success rate measures partial correctness by evaluating whether the temporal order of the predicted steps matches the ground truth steps. Two pointers are used: one for tracking the matched ground-truth steps (\texttt{ptr\_g}) and one for sweeping through the predicted steps (\texttt{ptr\_p}). The algorithm is as follows:

\begin{algorithm}[H]
\caption{SSR: Step Success Rate}
\begin{algorithmic}[1]
\STATE \textbf{Input:} Predicted steps $\mathcal{P}$, Ground truth steps $\mathcal{G}$
\STATE \textbf{Output:} SSR value
\STATE Initialize \texttt{MATCH} $\gets$ 0
\STATE Initialize \texttt{ptr\_g} $\gets$ 1
\STATE Initialize \texttt{ptr\_p} $\gets$ 1
\FOR{each predicted step $p \in \mathcal{P}$starting from \texttt{ptr\_p}} 
    \FOR{each ground truth step $g \in \mathcal{G}$ starting from \texttt{ptr\_g}}
        \IF{$p$ matches $g$}
            \STATE \texttt{MATCH} $\gets$ \texttt{MATCH} + 1
            \STATE \texttt{ptr\_g} $\gets$ \texttt{ptr\_g} + 1
            \STATE \textbf{break}
        \ENDIF
    \ENDFOR
\ENDFOR
\STATE \textbf{return} $\frac{\texttt{MATCH}}{||\mathcal{G}||}$
\end{algorithmic}
\end{algorithm}

For all three metrics, we report the average across all demonstrations as the performance of the models. Notably, we provide an automated evaluation script in our open-source code, which significantly reduces the need for time-consuming manual verification.

\begin{figure}[t]
    \centering
    \includegraphics[width=\linewidth]{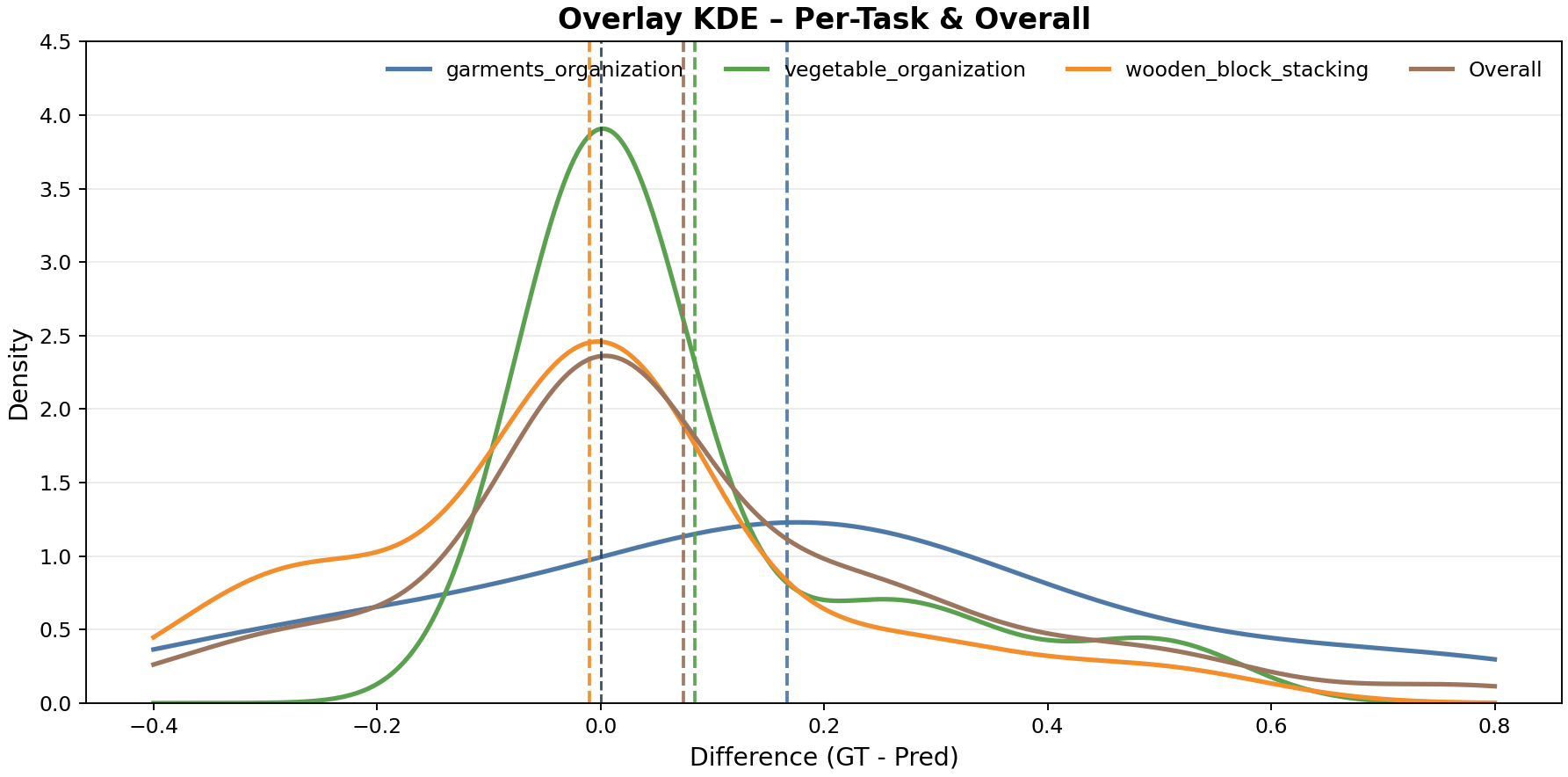}
    \vspace{-5mm}
    \caption{KDE plots of pointwise differences between the automated evaluation script and manually verified ground truth across three tasks and the overall. The vertical lines indicate the mean of the differences on different tasks.
    }
    \label{fig:kde}
\vspace{-5mm}
\end{figure}

\textbf{Automated Evaluation with LLM}. To validate the reliability of the automated evaluation script, we computed pointwise differences between the script-generated evaluation scores and the manually evaluated scores for each data point. We then visualized these differences using kernel density estimation (KDE) plots and reported their mean values in Fig~\ref{fig:kde}.
The mean differences for vegetable organization, garment organization, and wooden block organization are 0.084, 0.167, and 0.010, respectively, with an overall mean difference of 0.074 across all tasks. This indicates that, from a statistical perspective, the automated script’s measurements are close to those of manual evaluation. Garment organization exhibits a higher mean difference, likely due to the task’s higher complexity and the prevalence of ambiguous descriptions for visually similar garments, which are prone to mismatches with the ground truth.
In vegetable organization, garment organization, and wooden block stacking, the automated script produced results identical to manual evaluation for 27/38, 6/30, and 26/39 subtasks, respectively. These results collectively demonstrate that the automated evaluation script is statistically reliable and suitable for practical use.

\subsection{Real-world experiment}
The real-world experiment is conducted on a Universal Robots' UR10e cobot attached with a Robotiq 2F-85 gripper. We use an Intel RealSense 455 stereo camera to acquire depth information. The camera is mounted on a tripod standing across the robot and hand-eye calibrated in an eye-to-hand setup. Similar to \cite{liang2023code, yoneda2024statler}, we first use a segmentation model to segment all the objects of interest in the RGB images, and then we query the depth image with the same coordinates to acquire 3D information of the objects. The gripper is hard-coded to move horizontally above the object in 3D space and lower in the z-axis to grasp the object. Because the action primitives are hard-coded, it inevitably resulted in some failure cases due to lack of adaptability. For example, in the wood block stacking example, the dropping distance is often too high and would cause the wood block to bounce upon contact.

\end{document}